\pdfoutput=1
\documentclass[11pt]{article}
\usepackage[]{ACL2023}
\usepackage{makecell}
\usepackage{algorithmicx}
\usepackage{algorithm}
\usepackage{algpseudocode}
\usepackage{times}
\usepackage{latexsym}
\usepackage{xcolor}
\usepackage{soul}
\usepackage[T1]{fontenc}
\usepackage[utf8]{inputenc}
\usepackage{microtype}
\usepackage{inconsolata}
\usepackage{graphicx}
\usepackage{enumitem}
\usepackage{array}
\usepackage{url}
\usepackage{booktabs}
\usepackage{amssymb}
\usepackage{multirow}
\usepackage{pifont}
\usepackage{courier}
\usepackage{lipsum}
\usepackage{cleveref}
\usepackage{catchfile}
\usepackage{changes}
\newcommand{\cmark}{\ding{51}}
\newcommand{\xmark}{\ding{55}}

\title{Open-WikiTable: Dataset for Open Domain Question Answering\\
with Complex Reasoning over Table}

\author{Sunjun Kweon$^{1}$, Yeonsu Kwon$^{1}$, Seonhee Cho$^{1}$, \textbf{Yohan Jo$^{2}$\thanks{\hspace{0.2cm}This work is not associated with Amazon.}, Edward Choi$^{1}$} \\
  KAIST$^{1}$ , Amazon$^{2}$\\
  \texttt{\{sean0042, yeonsu.k, ehcho8564, edwardchoi\}@kaist.ac.kr} \\
  \texttt{\{jyoha\}@amazon.com} }

\begin{document}

\maketitle
\begin{abstract}

Despite recent interest in open domain question answering (ODQA) over tables, many studies still rely on datasets that are not truly optimal for the task with respect to utilizing \emph{structural nature of table}. These datasets assume answers reside as a single cell value and do not necessitate exploring over multiple cells such as aggregation, comparison, and sorting. Thus, we release Open-WikiTable, the first ODQA dataset that requires complex reasoning over tables. Open-WikiTable is built upon WikiSQL and WikiTableQuestions to be applicable in the open-domain setting. As each question is coupled with both textual answers and SQL queries, Open-WikiTable opens up a wide range of possibilities for future research, as both reader and parser methods can be applied. The dataset and code are publicly available\footnote{\url{https://github.com/sean0042/Open_WikiTable}}.

\end{abstract}

\section{Introduction}

Tables have played a prominent role as a source of knowledge in question answering (QA). They contain various types of data such as numeric, temporal, and textual information in a structured manner. Early table QA datasets \cite{pasupat2015compositional, zhong2017seq2sql, yu2018spider} have focused on complex questions that exploit the structure of tables via aggregation, comparison, or sorting.
However, these datasets assume that 
the relevant table is always given for each question \cite{kostic2021multi}, limiting their applicability in real-world scenarios.
For more practical use, recent works extend 
tableQA to the open-domain setting, where the evidence table should be retrieved solely from using the question itself. 

The first research of open-domain QA over tables is \citet{herzig2021open}. They released a dataset, NQ-table, by extracting questions from Natural Questions \cite{kwiatkowski2019natural} whose answers reside in a table.
All questions, however, are answered by extracting a single cell and do not necessitate any extensive reasoning across multiple cells.
It is also notable that 55\% of the evidence tables consist of only a single row, which has little structure. 

    \begin{figure}[t] 
    \begin{center}
    \includegraphics[width=1.0\linewidth]{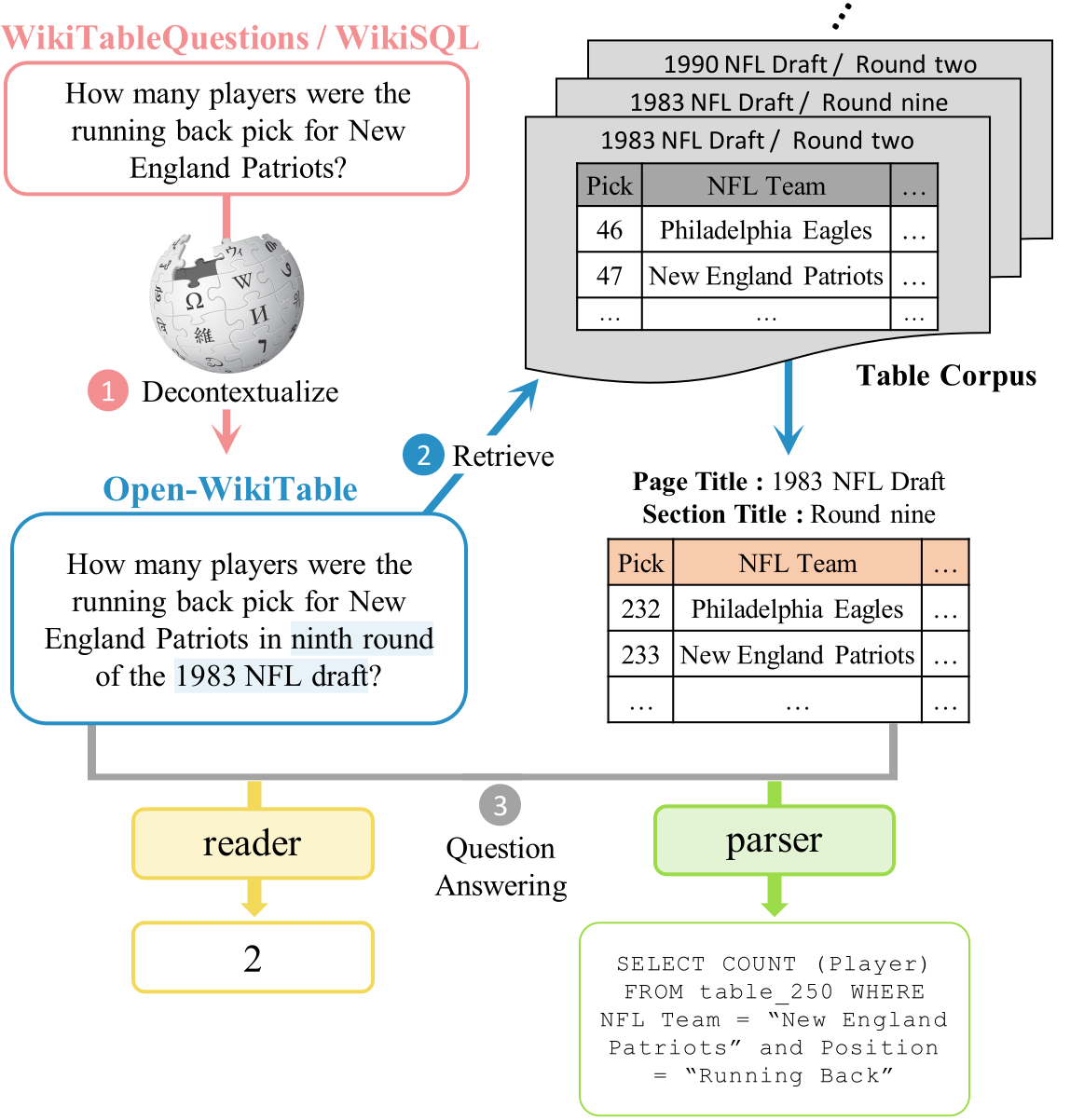}
    \end{center}
    \vspace{0mm}
    \caption{Open-WikiTable is built by revising WikiSQL and WikiTableQuestions. Through decontextualization, the question provides the necessary information to retrieve the grounding table. As the dataset is labeled with textual answers and SQL queries, both reader and parser approaches can be used.}
    \label{figure1}
    \vspace{-4mm}
    \end{figure}

    \begin{table*}[!htb]
    \centering
    \setlength{\heavyrulewidth}{1.5pt}
    \resizebox{2.0\columnwidth}{!}{
        \newcommand{\mytable}{\begin{tabular}{c|cccccc}
\toprule
\multirow{2}{*}{Datasets} & \multirow{2}{*}{\begin{tabular}[c]{@{}c@{}}Retrieval\\ Availability\end{tabular}} & \multirow{2}{*}{\begin{tabular}[c]{@{}c@{}}Complex\\ Reasoning\end{tabular}} & \multirow{2}{*}{\begin{tabular}[c]{@{}c@{}}\# of\\ QA pairs\end{tabular}} & \multirow{2}{*}{\begin{tabular}[c]{@{}c@{}} \# of\\ table corpus\end{tabular}} & \multirow{2}{*}{\begin{tabular}[c]{@{}c@{}}Answer\\ Annotation\end{tabular}} & \multirow{2}{*}{Reference} \\
&  &  &  &  &  &  \\ \hline
\textbf{WikiTableQuestions} &   \xmark  &  \cmark  & 22,033   & 2,108  & Text  &  \citet{pasupat2015compositional} \\
\textbf{WikiSQL}  &  \xmark  &  \cmark  & 80,654  & 24,241  & SQL  &  \citet{zhong2017seq2sql}  \\
\textbf{E2E-WTQ} &  \cmark  &  \xmark  & 1,216  & 2,108  & Text  &  \citet{pan2021cltr} \\
\textbf{E2E-GNQ} & \cmark & \xmark & 789  & 74,224 & Text & \citet{pan2021cltr} \\
\textbf{NQ-table}  &  \cmark  &  \xmark  & 11,628  & 169,898  & Text  & \citet{herzig2021open}  \\ \hline
\textbf{Open-WikiTable} & \cmark  &  \cmark  & 67,023 & 24,680 & Text, SQL & \\
\bottomrule
\end{tabular} }
            
\begin{tabular}{ll}
\Xhline{1.3pt} 
\textit{\textbf{Error type 1 (70\%)}}       & Ambiguity in the original question \\ 
\hline
\textit{\textbf{original}}          & Name the 2005 with 2007 of sf                                             \\
\textit{\textbf{de-contextualized}} & Name the 2005 with 2007's Doubles name of sf among Alicia Molik?          \\
\textit{\textbf{parapharsed}}       & Do you name the 2005s with 2007 doubles names from sf under Alicia Molik? \\
\Xhline{1.3pt}

\textit{\textbf{Error type 2 (15\%)}}       & Change of intent after decontextualizing \\ 
\hline
    \textit{\textbf{original}}          & how many total rounds did she fight in ?                                             \\
\textit{\textbf{de-contextualized}} & How many total rounds did she fight for Gina Carano?          \\
\textit{\textbf{parapharsed}}       & How many rounds did she fight for Gina Carano in total? \\ 
     \Xhline{1.3pt} 

\textit{\textbf{Error type 3 (15\%)}}       & Change of intent after paraphrasing \\ 
\hline
       \textit{\textbf{original}}          & Which Bask has an Indoor track of 0, and a Swimming of 5?                                             \\
\textit{\textbf{de-contextualized}} & Which bask has an indoor track of 0,and a swimming of 5 for horizon league's           \\
&women's sports championship totals?\\
\textit{\textbf{parapharsed}}       & Which pool has an indoor stretch of 0 and a swim of 5 for the total number  \\
&of women's athletic championships in the horizon league?\\ \Xhline{1.3pt}  
\end{tabular}

    }
    \caption{
    Comparison between table question answering datasets. 
    }
    \label{table1}
     \end{table*}

Another work for open-domain table QA is \citet{pan2021cltr}.
They presented E2E-WTQ and E2E-GNQ datasets, extensions of WikiTableQuestion \cite{pasupat2015compositional} and GNQtables \cite{shraga2020web}, to develop cell-level table retrieval models.
However, as they assume cell extraction for the table QA task and construct the datasets accordingly, E2E-WTQ and E2E-GNQ share the same limitation as the NQ-table; all answers are restricted to a single cell.
Another issue with these datasets is their small size, containing only around 1k examples in total.
This makes it challenging to train language models as there may not be enough data.


Given that there is currently no dataset that fully considers the structural property in the open-domain setting, we present \textbf{Open-WikiTable}.
It extends WikiSQL and WikiTableQuestions to be more applicable in the open-domain setting.
Open-WikiTable is a large-scale dataset composed of 67,023 questions with a corpus of 24,680 tables. The key features of our dataset are listed below.

\begin{itemize}[leftmargin=3mm]
\item First, nearly 40\% of the questions require advanced reasoning skills beyond simple filtering and cell selection. The model should utilize operations such as aggregating, sorting, and comparing multiple cells to derive an accurate answer.
\end{itemize}

\begin{itemize}[leftmargin=3mm]
\item Second, all questions are carefully designed for the retrieval task in the open-domain setting.
We manually re-annotated 6,609 table descriptions (i.e. page title, section title, caption), then added them to the original question to ensure that questions convey sufficient context to specify the relevant table.
\end{itemize}

\begin{itemize}[leftmargin=3mm]
\item Third, questions are paraphrased to reduce high word overlap between the question and the grounding table.
It reflects a tendency in the open-domain setting where questions are often phrased in diverse styles, and terms in the questions may be different from those in the table.
\end{itemize}

\begin{itemize}[leftmargin=3mm]
\item Lastly, every question in the dataset is labeled with both textual answers and SQL queries. This provides an opportunity to train and evaluate models with both common table QA techniques, \emph{Reader} and \emph{Parser} in parallel.
\end{itemize}

In this work, we thoroughly explain the data construction process of Open-WikiTable. We perform open domain question answering by incorporating a retrieval task with QA over tables (see Figure \ref{figure1}). Then, 
we evaluate the performance of the retriever and the QA models with both reader and parser approaches.

\section{Data Construction} \label{section1}

\textbf{Open-WikiTable} is built upon two closed-domain table QA datasets - WikiSQL \cite{zhong2017seq2sql} and WikiTableQuestions \cite{pasupat2015compositional}. WikiSQL is a large-scale text-to-SQL dataset but is composed of relatively simple questions since they are constructed from limited SQL templates. WikiTableQuestions contains more complex questions involving superlative or arithmetic expressions but only provides textual form answers. \citet{shi2020potential} further annotated SQLs for a subset of WikiTableQuestions. By utilizing these datasets, we aim to create a diverse and intricate set of questions, with each question annotated with both a textual and logical form answer.

Although the questions in WikiSQL and WikiTableQuestions require more advanced table reasoning than those of existing open-domain table QA datasets (See Table \ref{table1}), they possess two problems to be directly used in the open domain setting.
First, questions are not specific enough to retrieve relevant tables.
Second, questions have high word overlaps with table contents which are unrealistic in the open-domain setting where the question can be expressed in lexically diverse forms. 
We resolve the first issue via decontextualization (\ref{Decontextualization}) and the second issue via paraphrasing (\ref{Paraphrase}), as elaborated in the following sections.

\subsection{Decontextualization} \label{Decontextualization}

Our goal is to decontextualize questions, that is, adding enough context about relevant tables to each question so that retrievers can find the relevant tables \cite{chen2020open, choi2021decontextualization}. However, the obstacle here is that a significant portion of table descriptions provided by WikiSQL and WikiTableQuestions were either missing or not specifically described to distinguish between tables. In this case, decontextualized questions still cannot point out the exact grounding tables (appendix \ref{Appendix A.1}). Therefore, we resolved this issue by comparing 6,609 problematic tables with the corresponding Wikipedia article and re-annotating table descriptions. The resulting table corpus of Open-WikiTable has 24,680 tables, all of which have distinct descriptions.


Next, the questions were decontextualized with the re-annotated table descriptions. All table descriptions necessary for the retrieval of the grounding table were incorporated into each question.
We transformed the questions by utilizing GPT-J, a language model from Eleuther AI.
In order to ensure that the generated question accurately reflects the original intention, we decontextualized the questions by maintaining the form of the original question while incorporating table descriptions only as adverbs, as exemplified in Appendix \ref{Appendix A.2}.
The generated questions were accepted only if all key entities (i.e. referred column names and condition values) of the original question and added table descriptions were preserved.
If not, we repeatedly generated new samples until accepted.

\subsection{Paraphrase} \label{Paraphrase}

Although the decontextualization process ensures the questions are suitable for table retrieval, it is quite easy to retrieve the grounding table due to a high degree of word overlap in the question and the table contents. To address this issue, we further paraphrased the questions via back-translation \cite{prabhumoye2018style}.
We utilized English-German-English translation using Google Translate API. To inspect whether the degree of word overlap has decreased, we measure the average BLEU score between the question and grounding table contents. It has dropped after paraphrase, from $7.28\times10^{-2}$ to $6.56\times10^{-2}$.
It is also notable that the variance of word distribution in the questions has increased from $2.3\times10^{5}$ to $3.1\times10^{5}$ through paraphrasing.

\subsection{Quality Check}
We then review the questions to ensure their quality as the final step. Authors manually reviewed 10k randomly selected questions, according to the following standards: 1) The intent of the original question should not be altered during any stage of the data construction process. 2) Every information added through the decontextualization process should be preserved after paraphrasing. It turned out that 7.9\% of 10k randomly selected samples did not meet our criteria.
Within the 7.9\% error rate, we discovered that 70\% of these errors were due to the ambiguity of the original question. As a result, errors stemming from our decontextualization and paraphrasing processes account for 2.3\% of the 10,000 random samples. The final test set, however, is composed only of the accepted samples during the quality review  to ensure the integrity in the evaluation of the model performance.  Error examples are reported in appendix \ref{Appendix A.3}.

\subsection{Data Statistics}
As part of our dataset preparation process, we partitioned the entire dataset into train, validation, and test sets, with a ratio of 8:1:1. Consequently, the test set comprised 6,602 instances, as shown in Table \ref{statistics}. It is important to note that during this partitioning process, we ensured that each subset do not share any tables, enabling us to evaluate the generalizability of the models to previously unseen tables.

    \begin{table}[H] 
    \centering
    \setlength{\heavyrulewidth}{1.5pt}
    \resizebox{1.0\columnwidth}{!}{
    
\begin{tabular}{|c|cccc|}
\hline
             & \textbf{Train}  & \textbf{Valid} & \textbf{Test}  & \textbf{Total}  \\ \hline
\textbf{\# of questions} & 53,819 & 6,602 & 6,602 & 67,023 \\
\textbf{\# of tables }   & 17,275 & 2,139 & 2,262 & 21,676 \\ \hline
\textbf{corpus size}  & \multicolumn{4}{c|}{24,680}     \\ \hline
\end{tabular}
    }
    \caption{Statistics of Open-WikiTable}
    \label{statistics}
    \end{table}

\section{Experiments}
First of all, we split tables into segments so that models can handle long tables within the limited input sequence length. Inspired by \citet{karpukhin2020dpr}, tables are split row-wise into 100-word chunks.
Around 52\% of tables in our corpus are split into multiple chunks, which resulted in a total of 54,282 table segments. 
For the retrieval task, each table is flattened and appended with the table descriptions, and then fed to a retriever. When a grounding table is split into multiple tables, all table segments that are relevant to an answer should be retrieved. Then, we perform end-to-end table QA where the model should answer the question given retrieved tables.
More details about experimental settings are in Appendix \ref{Appendix B}.

\subsection{Retrieval}

    \begin{table}[t]
    \centering
    \setlength{\heavyrulewidth}{1.5pt}
    \resizebox{1.0\columnwidth}{!}{
        \newcommand{\mytable}{\begin{tabular}{cc|c|ccc}
\toprule
\multicolumn{2}{c|}{\textbf{Encoder}} & \multirow{2}{*}{\textbf{Data}} & \multirow{2}{*}{\textbf{k=5}} & \multirow{2}{*}{\textbf{k=10}} & \multirow{2}{*}{\textbf{k=20}} \\
Text & Table & & & & \\ 
\hline
\multicolumn{2}{c|}{\multirow{3}{*}{BM25}}     & Original              & 6.6                  & 8.0                   & 10.3                  \\
\multicolumn{2}{c|}{}                          & Decontextualized      & 45.5                 & 52.9                  & 59.7                  \\
\multicolumn{2}{c|}{}                          & Paraphrased           & 42.2                 & 48.9                  & 56.1                  \\ 
\hline
\multirow{3}{*}{BERT} & \multirow{3}{*}{BERT}  & Original              & 25.0                 & 34.1                  & 45.1                  \\
                      &                        & Decontextualized      & 91.6                 & 96.0                  & 97.8                  \\
                      &                        & Paraphrased           & 89.5                 & 95.0                  & 97.3                  \\ 
\hline
\multirow{3}{*}{BERT} & \multirow{3}{*}{TAPAS} & Original              & 19.4                 & 28.1                  & 38.5                  \\
                      &                        & Decontextualized      & 88.2                 & 94.5                  & 97.3                  \\
                      &                        & Paraphrased           & 84.0                 & 91.4                  & 95.6                  \\
\bottomrule
\end{tabular}}
            
\begin{tabular}{ll}
\Xhline{1.3pt} 
\textit{\textbf{Error type 1 (70\%)}}       & Ambiguity in the original question \\ 
\hline
\textit{\textbf{original}}          & Name the 2005 with 2007 of sf                                             \\
\textit{\textbf{de-contextualized}} & Name the 2005 with 2007's Doubles name of sf among Alicia Molik?          \\
\textit{\textbf{parapharsed}}       & Do you name the 2005s with 2007 doubles names from sf under Alicia Molik? \\
\Xhline{1.3pt}

\textit{\textbf{Error type 2 (15\%)}}       & Change of intent after decontextualizing \\ 
\hline
    \textit{\textbf{original}}          & how many total rounds did she fight in ?                                             \\
\textit{\textbf{de-contextualized}} & How many total rounds did she fight for Gina Carano?          \\
\textit{\textbf{parapharsed}}       & How many rounds did she fight for Gina Carano in total? \\ 
     \Xhline{1.3pt} 

\textit{\textbf{Error type 3 (15\%)}}       & Change of intent after paraphrasing \\ 
\hline
       \textit{\textbf{original}}          & Which Bask has an Indoor track of 0, and a Swimming of 5?                                             \\
\textit{\textbf{de-contextualized}} & Which bask has an indoor track of 0,and a swimming of 5 for horizon league's           \\
&women's sports championship totals?\\
\textit{\textbf{parapharsed}}       & Which pool has an indoor stretch of 0 and a swim of 5 for the total number  \\
&of women's athletic championships in the horizon league?\\ \Xhline{1.3pt}  
\end{tabular}

    }
    \caption{
    Top-k table retrieval accuracy on three different construction stages of Open-WikiTable's validation set.
    }
    \vspace{-4mm}
    \label {table2}
    \end{table}

\paragraph{Experimental Setup}
We employ the BM25 algorithm \cite{robertson2009probabilistic} for the sparse search. For the dense search, we utilize a dual-encoder approach: BERT \cite{devlin2018bert} and TAPAS \cite{herzig2020tapas} for the table encoder and BERT for the question encoder. They are trained to maximize the inner product between the question and table embeddings. 
The performance of the retriever is measured at different top-\emph{k} retrieval accuracy, where we use 5, 10, and 20 for \emph{k}. To analyze the effect of each data construction process on the retrieval task, we experiment with three different types of questions: original question, decontextualized question, and paraphrased question. The result is shown in Table \ref{table2}.

\paragraph{Result}

Our experiments demonstrate that decontextualizing led to improved performance in all experiments. 
This suggests that the original questions are not sufficient for table retrieval and decontextualization dramatically alleviates this problem.
However, the result also implies that table retrieval becomes too easy as the information is added directly to the question without any syntactic or semantic changes. 
This tendency is mitigated after paraphrasing, which led to a performance drop for all retrievers. Specifically, BM25 had the largest performance drop, while the methods utilizing language models had relatively smaller drops, demonstrating their robustness against linguistic variation. These results suggest that word overlap between questions and tables is reduced and advanced semantic understanding is required.
Additionally, when comparing the performance of BERT and TAPAS table encoders, retrieval performed better with BERT for all three types of data. As previously demonstrated by \citet{wang2022table} in the case of NQ-table, table retrieval does not necessarily require a table-specific model, a conclusion reconfirmed by Open-WikiTable.

\subsection{End-to-End Table QA}

    \begin{table}[t]
    \centering
    \setlength{\heavyrulewidth}{1.5pt}
    \resizebox{1.0\columnwidth}{!}{
        \newcommand{\mytable}{
\begin{tabular}{c|ccc|ccc}
\toprule
\multirow{2}{*}{\textbf{Method}} & \multicolumn{3}{c|}{\textbf{Validation}} & \multicolumn{3}{c}{\textbf{Test}} \\
                       & k=5     & k=10    & k=20   & k=5    & k=10   & k=20   \\
\midrule
Reader                 & 55.1    & 62.7    & 65.0   & 57.5   & 64.5   & 65.2   \\
\midrule
Parser                 & 63.3    & 66.0    & \textbf{67.0}   & 65.2   & 67.1   & \textbf{67.9}  \\
\bottomrule

\end{tabular}}
            
\begin{tabular}{ll}
\Xhline{1.3pt} 
\textit{\textbf{Error type 1 (70\%)}}       & Ambiguity in the original question \\ 
\hline
\textit{\textbf{original}}          & Name the 2005 with 2007 of sf                                             \\
\textit{\textbf{de-contextualized}} & Name the 2005 with 2007's Doubles name of sf among Alicia Molik?          \\
\textit{\textbf{parapharsed}}       & Do you name the 2005s with 2007 doubles names from sf under Alicia Molik? \\
\Xhline{1.3pt}

\textit{\textbf{Error type 2 (15\%)}}       & Change of intent after decontextualizing \\ 
\hline
    \textit{\textbf{original}}          & how many total rounds did she fight in ?                                             \\
\textit{\textbf{de-contextualized}} & How many total rounds did she fight for Gina Carano?          \\
\textit{\textbf{parapharsed}}       & How many rounds did she fight for Gina Carano in total? \\ 
     \Xhline{1.3pt} 

\textit{\textbf{Error type 3 (15\%)}}       & Change of intent after paraphrasing \\ 
\hline
       \textit{\textbf{original}}          & Which Bask has an Indoor track of 0, and a Swimming of 5?                                             \\
\textit{\textbf{de-contextualized}} & Which bask has an indoor track of 0,and a swimming of 5 for horizon league's           \\
&women's sports championship totals?\\
\textit{\textbf{parapharsed}}       & Which pool has an indoor stretch of 0 and a swim of 5 for the total number  \\
&of women's athletic championships in the horizon league?\\ \Xhline{1.3pt}  
\end{tabular}

    }
    \caption{
    Comparison of the end-to-end reader and parser's exact match (EM) score, where \emph{k} represents the number of tables retrieved.
    }
    \label{table3}
    \end{table}

\paragraph{Experimental Setup}
We experiment with two different methods: reader and parser. Conventionally, the parser only utilizes the table schema rather than the entire contents, as the question typically specifies the exact table value. However, in Open-WikiTable, the values are often paraphrased, requiring the parser to extract the exact value from the table contents (See Appendix \ref{Appendix C}). 

For end-to-end question answering, we adopt the retriever that yielded the highest performance in the previous experiment. The question and retrieved tables are concatenated and fed to QA models. Both reader and parser are implemented with the fusion-in-decoder architecture \cite{izacard2020leveraging} and the T5-\emph{base} language model \cite{raffel2020exploring}. We use the exact match accuracy (EM) for the evaluation metric. 
For the parser, EM is computed on the execution result of generated SQLs, as they can be expressed in a diverse form.

\paragraph{Result}

Table \ref{table3} summarizes validation and test results for end-to-end QA. As the retrieval performance improves with increased \emph{k}, QA models, which rely on the retrieved tables, accordingly show consistent performance improvement with larger \emph{k}. 
However, regardless of the number of \emph{k}, the parser model outperforms the reader model. This performance gap is most significant with small \emph{k}, and decreases as \emph{k} grows. 
We posit that this is due to the difference in the minimum amount of table segments that the reader and parser must refer to create an accurate answer.
The parser model can generate a correct SQL query even when all segments of a table are not retrieved, as long as any of the retrieved splits possess all necessary cell values. On the contrary, the reader model should refer to every relevant split to derive a correct answer.

    \begin{table}[t]
    \centering
    \setlength{\heavyrulewidth}{1.5pt}
    \resizebox{1\columnwidth}{!}{
        \newcommand{\mytable}{\begin{tabular}{cc|cc|c}
\Xhline{1.3pt}
\multicolumn{2}{c|}{\textbf{Category}}                              & \multirow{2}{*}{\textbf{Reader}} & \multirow{2}{*}{\textbf{Parser}} & \multirow{2}{*}{\textbf{\# questions}} \\ \cline{1-2}
\multicolumn{1}{c|}{\textbf{Table-split}}             & \textbf{Complexity} &                         &                         &                               \\ \hline
\multicolumn{1}{c|}{\multirow{2}{*}{Single}} & Easy        & 74.0                    & \textbf{82.8}                    & 1,574                         \\
\multicolumn{1}{c|}{}                        & Hard        & \textbf{62.9}                    & 51.2                    & 1,794                         \\ \hline
\multicolumn{1}{c|}{\multirow{2}{*}{Multi}}  & Easy        & 70.5                    & \textbf{82.5}                    & 1,520                         \\
\multicolumn{1}{c|}{}                        & Hard        & 56.0                    & \textbf{58.8}                    & 1,714  
\\
\Xhline{1.3pt}
\end{tabular}}
            
\begin{tabular}{ll}
\Xhline{1.3pt} 
\textit{\textbf{Error type 1 (70\%)}}       & Ambiguity in the original question \\ 
\hline
\textit{\textbf{original}}          & Name the 2005 with 2007 of sf                                             \\
\textit{\textbf{de-contextualized}} & Name the 2005 with 2007's Doubles name of sf among Alicia Molik?          \\
\textit{\textbf{parapharsed}}       & Do you name the 2005s with 2007 doubles names from sf under Alicia Molik? \\
\Xhline{1.3pt}

\textit{\textbf{Error type 2 (15\%)}}       & Change of intent after decontextualizing \\ 
\hline
    \textit{\textbf{original}}          & how many total rounds did she fight in ?                                             \\
\textit{\textbf{de-contextualized}} & How many total rounds did she fight for Gina Carano?          \\
\textit{\textbf{parapharsed}}       & How many rounds did she fight for Gina Carano in total? \\ 
     \Xhline{1.3pt} 

\textit{\textbf{Error type 3 (15\%)}}       & Change of intent after paraphrasing \\ 
\hline
       \textit{\textbf{original}}          & Which Bask has an Indoor track of 0, and a Swimming of 5?                                             \\
\textit{\textbf{de-contextualized}} & Which bask has an indoor track of 0,and a swimming of 5 for horizon league's           \\
&women's sports championship totals?\\
\textit{\textbf{parapharsed}}       & Which pool has an indoor stretch of 0 and a swim of 5 for the total number  \\
&of women's athletic championships in the horizon league?\\ \Xhline{1.3pt}  
\end{tabular}

    }
    \caption{
Exact match scores on Open-WikiTable's test set with \emph{k}=20, where we categorize questions by their complexity and whether the grounding table is split.
    }
    \label{table4}
    \end{table}

For more detailed analysis, we categorize questions into easy or hard based on if the answer is derived from a single cell value, and into single-table or multi-table based on if the grounding table is split.
The results are shown in Table \ref{table4}. The parser outperforms the reader when the grounding table is split into multiple segments, regardless of question complexity, which aligns with the previous analysis.
It is notable that the parser shows inferior or comparable performance to the reader for hard questions.
We believe this is due to the relative size between WikiSQL (\textit{i.e.} mostly easy) and WikiTableQuestions (\textit{i.e.} mostly hard), and that the parser has a limited opportunity to understand the diversity of complex SQL queries.


\section{Conclusion}

We present Open-WikiTable, the first ODQA dataset that requires complex reasoning over Wikipedia tables. The dataset is constructed by revising WikiTableQuestions and WikiSQL to be fully functional in the open-domain setting through decontextualization and paraphrasing. The dataset provides both textual and logical form answers for each question so that end-to-end reader and parser models can be trained. 
We hope that Open-WikiTable can provide new opportunities for future research such as investigating the effectiveness of leveraging both reader and parser approaches in the retrieval and generation phase.

\section*{Limitations}
Although we carefully designed Open-WikiTable for complex open-domain table QA, there are some limitations since it is based on the existing datasets. First, ambiguous or erroneous samples from the original WikiSQL or WikiTableQuestions dataset may still lie in our training and validation set. As we mentioned in Section 3.2, most of the equivocal samples were attributed to the ambiguity of the original question and excluded from the test set, but not removed. Second, unlike semantic coverage of the questions is extended by decontextualization and paraphrasing, the coverage of the question remains in that the answer and logic to derive the answer in each question is the same. Still, Open-WikiTable demonstrates the potential for further research on open-domain QA over the table.

\section*{Ethics Statement}
No ethics concerned with our work.


\bibliography{custom}
\bibliographystyle{acl_natbib}

\clearpage

\appendix

\section{Data Construction Details} \label{Appendix A}

    \begin{figure*}[!htpb] 
    \includegraphics[width=1.0\linewidth]{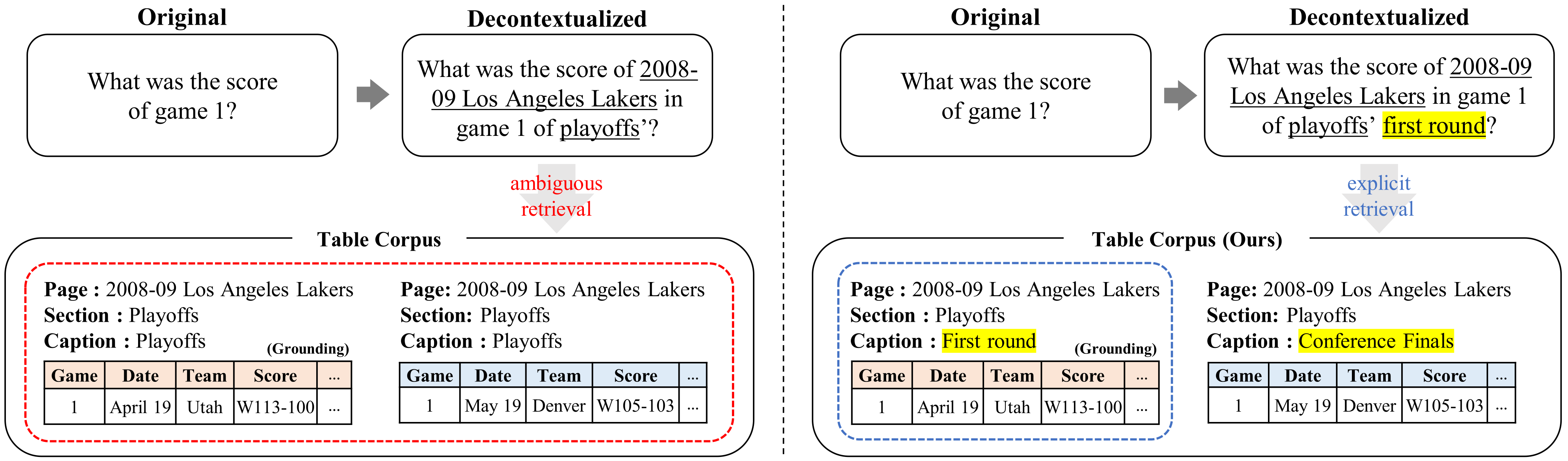}
    \vspace{-3mm}
    \caption{Comparison on retrieval of decontextualized question using WikiSQL(left) and Open-WikiTable(right). The table description of WikiSQL is insufficient to pinpoint the grounding table, even after the decontextualizing process. In contrast, the descriptions of Open-WikiTable effectively address the issue by re-annotation.}
    \label{figure2}
    \vspace{-4mm}
    \end{figure*}

\subsection{Table Descriptions Re-Annotation} \label{Appendix A.1}

Figure \ref{figure2} illustrates the indistinguishable annotation of the table corpus in WikiSQL and WikiTableQuestions, leading to ambiguity in the decontextualized questions. The figure on the right shows how the problem is solved by re-annotating the table descriptions.

\subsection{Construction Details} \label{Appendix A.2}

The prompt used by GPT-J for decontextualization can be found in Table \ref{table5}. Table \ref{table6} shows examples of each step in the process of creating the Open-WikiTable.

\subsection{Error Analysis} \label{Appendix A.3}
Upon closer examination of the 7.9\% error on generated Open-WikiSQL, we find that 70\% of the errors were the results of ambiguity in the original questions, which was propagated over during the data construction process. The percentage of errors by the decontextualization process and paraphrasing process was 15\% respectively. In Table \ref{table7}, we provide examples for each type of error encountered.



    

\section{Experimental Setup} \label{Appendix B}

\subsection{Flattened Table Format}
In order to present the table as passages, we flattened the table and added table descriptions with the help of special tokens. For example, \vspace{3mm}

\resizebox{0.9\columnwidth}{!}{
    \begin{tabular}{lcccc}
    \Xhline{1.3pt}
\multicolumn{5}{l}{\textbf{Page title} : 2008-09 Los Angeles Lakers}          \\
\multicolumn{5}{l}{\textbf{Section title} : Playoffs}                         \\
\multicolumn{5}{l}{\textbf{Caption} : First round}                            \\ 
\multicolumn{5}{l}{\textbf{Table ID} : table\_132938\_29}                            \\ 
\multicolumn{5}{l}{}                                                 \\ \hline
\multicolumn{1}{c}{\textbf{Game}} & \textbf{Date}     & \textbf{Team} & \textbf{Score}     & \textbf{High points} \\ \hline
\multicolumn{1}{c}{1}    & April 19 & Utah & W 113-100 & Kobe Bryant \\
\multicolumn{1}{c}{2}    & April 21 & Utah & W 119-109 & Kobe Bryant \\
\Xhline{1.3pt}
\end{tabular}
} \\ \\ \\   is flattened as

\begin{changemargin}{0.5cm}{0.5cm} 
\emph{[Page Title] 2008-09 Los Angeles Lakers [Section Title] Playoffs [Caption] First round [table\_id] table\_132938\_29 [Header] Game [SEP] Date [SEP] Team [SEP] Score [SEP] High points [Rows] [Row] 1 [SEP] April 19 [SEP] Utah [SEP] W 113-100 [SEP] Kobe Bryant [Row] 2 [SEP] April 21 [SEP] Utah [SEP] W 119-109 [SEP] Kobe Bryant}
\end{changemargin}

\subsection{Hyperparameters}

All experiments were on 8 NVIDIA A6000 48G GPUs. For the retrieval models, we use a batch size of 64, with a learning rate of 1.0 e-5 using Adam and linear scheduling with a warm-up. The in-batch negative technique was utilized to train the retriever. We evaluated every 500 steps and used early stopping with patience 5. For the question-answering module, we use batch size 8 for \emph{k} = 5, 10 and batch size 4 for \emph{k} = 20. The rest of the hyperparameters goes the same with the retriever.

\section{Open-WikiTable with Parser} \label{Appendix C} 
In the open-domain scenario, where the table is not specified a priori, questions may not contain the exact cell value to generate SQLs.
As shown below, it is necessary to refer to the grounding table and use the exact value to generate the correct SQL.

    \begin{table}[H] 
    \centering
    \setlength{\heavyrulewidth}{1.5pt}
    \resizebox{1.0\columnwidth}{!}{
\begin{tabular}{ll}
\Xhline{1.3pt}
\textbf{Question} & \begin{tabular}[c]{@{}l@{}}What is Born-Deceased if the term \\ of office is {\color[HTML]{9D0C0C}December 4, 1941} in \\ the list of Prime Ministers of Albania\end{tabular} \\
\hline
\textbf{SQL}      & \begin{tabular}[c]{@{}l@{}}SELECT Born\_Died From table\_2\\ WHERE Term\_start = \\ "{\color[HTML]{00009B}4 December 1941}"\end{tabular}                                      \\ \Xhline{1.3pt}
\textbf{Question} & \begin{tabular}[c]{@{}l@{}}In the Gothic-Germanic strong \\ verb, which part 2 has a verb \\ meaning {\color[HTML]{9D0C0C}to jump}?\end{tabular}                              \\ \hline
\textbf{SQL}      & \begin{tabular}[c]{@{}l@{}}SELECT Part\_2 FROM table\_3 \\ WHERE Verb\_meaning = \\ "{\color[HTML]{00009B}to leap}"\end{tabular}                                              \\ \Xhline{1.3pt}
\end{tabular} }
    \end{table}

 \begin{table*}[!ht]
    \centering
  
    \setlength{\heavyrulewidth}{1.5pt}
    \resizebox{\textwidth}{!}{
        \newcommand{\mytable}{\begin{tabular}{l}
\Xhline{1.5pt} 
Page Title : Wake Forest Demon Deacons football, 1980–89 \\ 
Section Title : Schedule\\
Caption : 1987\\    
Question : Who was the opponent when the result was L 0-14?    \\
What is converted question using given information?\\
In 1987's schedule, who was the opponent of Wake Forest Demon Deacons when the result was L 0-14?\\
...\\
Page Title : Toronto Raptors all-time roster \\ 
Section Title : A\\
Caption : A\\    
Question : What is order S24's LNER 1946 number?    \\
What is converted question using given information?\\
Considering the history of GER Class R24, what is order S24's LNER 1946 number?\\
...\\
Page Title : GER Class R24 \\ 
Section Title : History\\
Caption : Table of orders and numbers\\    
Question : What is order S24's LNER 1946 number?    \\
What is converted question using given information?\\
Considering the history of GER Class R24, what is order S24's LNER 1946 number?\\
...\\
Page Title : 2006–07 Toronto Raptors season \\ 
Section Title : Game log\\
Caption : February\\    
Question : Who had the highest number of rebounds on February 14?    \\
What is converted question using given information?\\
From 2006-07 Toronto Raptors' game log, who had the highest number of rebounds on February 14?\\
...\\
Page Title : Toronto Raptors all-time roster \\ 
Section Title : O\\
Caption : O\\    
Question : Which school was in Toronto in 2001-02?    \\
What is converted question using given information?\\
Which school was in Toronto in 2001-02 from Toronto Raptors all-time roster O?\\
...\\
Page Title : Stozhary \\ 
Section Title : Stozhary 2003 Prize-Winners\\
Caption : Stozhary 2003 Prize-Winners\\    
Question : What actor was nominted for an award in the film Anastasiya Slutskaya? \\
What is converted question using given information?\\
For 2003 Stozhary prize winners, what actor was nominted for an award in the film Anastasiya Slutskaya?\\
...\\
Page Title : 1985 New England Patriots season \\ 
Section Title : Regular season\\
Caption : Regular season\\    
Question : How many weeks are there?    \\
What is converted question using given information?\\
In 1985 New England Patriots season, how many weeks were there for regular season?\\
...\\
Page Title : Friday Night Lights (U.S. ratings) \\ 
Section Title : Weekly ratings\\
Caption : Season 1\\    
Question : What is the rank number that aired october 26, 2007?    \\
What is converted question using given information?\\
What is the rank number of Friday Night Lights Season 1's weekly ratings that aired october 26, 2007?\\
\Xhline{1.5pt} 
\end{tabular}}
            
\begin{tabular}{ll}
\Xhline{1.3pt} 
\textit{\textbf{Error type 1 (70\%)}}       & Ambiguity in the original question \\ 
\hline
\textit{\textbf{original}}          & Name the 2005 with 2007 of sf                                             \\
\textit{\textbf{de-contextualized}} & Name the 2005 with 2007's Doubles name of sf among Alicia Molik?          \\
\textit{\textbf{parapharsed}}       & Do you name the 2005s with 2007 doubles names from sf under Alicia Molik? \\
\Xhline{1.3pt}

\textit{\textbf{Error type 2 (15\%)}}       & Change of intent after decontextualizing \\ 
\hline
    \textit{\textbf{original}}          & how many total rounds did she fight in ?                                             \\
\textit{\textbf{de-contextualized}} & How many total rounds did she fight for Gina Carano?          \\
\textit{\textbf{parapharsed}}       & How many rounds did she fight for Gina Carano in total? \\ 
     \Xhline{1.3pt} 

\textit{\textbf{Error type 3 (15\%)}}       & Change of intent after paraphrasing \\ 
\hline
       \textit{\textbf{original}}          & Which Bask has an Indoor track of 0, and a Swimming of 5?                                             \\
\textit{\textbf{de-contextualized}} & Which bask has an indoor track of 0,and a swimming of 5 for horizon league's           \\
&women's sports championship totals?\\
\textit{\textbf{parapharsed}}       & Which pool has an indoor stretch of 0 and a swim of 5 for the total number  \\
&of women's athletic championships in the horizon league?\\ \Xhline{1.3pt}  
\end{tabular}

    }
    \caption{The prompt used for GPT-J when decontextualizing the question}
    \label{table5}
    \end{table*}
    
     \begin{table*}[!ht]
    \centering
    \setlength{\heavyrulewidth}{1.5pt}
    \resizebox{2.0\columnwidth}{!}{
        \newcommand{\mytable}{

\begin{tabular}{ll}
\Xhline{1.5pt} 


\textit{\textbf{original}   }                & The nhl team new york islanders is what nationality?
 \\
{\textit{\textbf{de-contexualized}}} & {\color[HTML]{00009B} The NHL team New York Islanders in what nationality 1994 } \\
                           &  {\color[HTML]{00009B} NHL Entry Draft's Round one? }                 \\
{\textit{\textbf{parapharsed}} }     &  {\color[HTML]{9D0C0C} What nationality is the NHL team New York Islanders}                                                   \\
                           & {\color[HTML]{9D0C0C} in the first round of the 1994 NHL Entry Draft? }                                                  \\ 

\hline

\textit{\textbf{original}   }                & What is the maximum starts that result in an average finish of 16.5?
 \\
{\textit{\textbf{de-contexualized}}} & {\color[HTML]{00009B}  What is the maximum starts that result in an average finish of 16.5 } \\
                           &  {\color[HTML]{00009B} for NASCAR Nationwide Series' Chad Little?
 }                 \\
{\textit{\textbf{parapharsed}} }     &  {\color[HTML]{9D0C0C} What are the maximum starts that result in a 16.5 average finish}                                                   \\
                           & {\color[HTML]{9D0C0C} for NASCAR Nationwide Series' Chad Little?  }                                                  \\ 

\hline


                          
         

\textit{\textbf{original}   }                &If the population is 2188, what was the median household income?
 \\
{\textit{\textbf{de-contexualized}}} & {\color[HTML]{00009B} If the population is 2188 in Ohio locations ranked by per capita income, } \\
                           &  {\color[HTML]{00009B} what was the median household income?
 }                 \\
{\textit{\textbf{parapharsed}} }     &  {\color[HTML]{9D0C0C} If Ohio's population is 2,188 ranked by per capita income,}                                                   \\
                           & {\color[HTML]{9D0C0C}what was the median household income?  }    \\    
\hline

\textit{\textbf{original}   }                &What values of HDTV correspond to  n° of 862?
 \\
{\textit{\textbf{de-contexualized}}} & {\color[HTML]{00009B} From the list of television in Italy's Shopping section,  } \\
                           &  {\color[HTML]{00009B}what values of HDTV correspond to n° of 862?
 }                 \\
{\textit{\textbf{parapharsed}} }     &  {\color[HTML]{9D0C0C}Which HDTV values correspond to the number 862 in the TV list}                                                   \\
                           & {\color[HTML]{9D0C0C} in the Italian shopping area? }

\\
\hline

\textit{\textbf{original}   }                &How many stories is the torre reforma building?
 \\
{\textit{\textbf{de-contexualized}}} & {\color[HTML]{00009B} How many stories is the torre reforma building from the list of } \\
                           &  {\color[HTML]{00009B}tallest buildings in Mexico's Under construction?
 }                 \\
{\textit{\textbf{parapharsed}} }     &  {\color[HTML]{9D0C0C}From the list of tallest buildings under construction in Mexico,}                                                   \\
                           & {\color[HTML]{9D0C0C}how many floors does the Torre Reforma building have? }

\\
\hline

\textit{\textbf{original}   }                &How many teams have a head coach named mahdi ali?

 \\
{\textit{\textbf{de-contexualized}}} & {\color[HTML]{00009B} How many teams has a head coach named mahdi ali } \\
                           &  {\color[HTML]{00009B} among 2010–11 UAE Pro-League?
 }                 \\
{\textit{\textbf{parapharsed}} }     &  {\color[HTML]{9D0C0C}How many teams in UAE Pro-League 2010-11 have a}                                                   \\
                           & {\color[HTML]{9D0C0C} head coach named Mahdi Ali?}

\\
\hline

\textit{\textbf{original}   }                &Which Member has an Electorate of southern melbourne?

 \\
{\textit{\textbf{de-contexualized}}} & {\color[HTML]{00009B} Which Member has an Electorate of southern melbourne among Members of } \\
                           &  {\color[HTML]{00009B} the Australian House of Representatives, 1903–1906?
 }                 \\
{\textit{\textbf{parapharsed}} }     &  {\color[HTML]{9D0C0C}Among the Members of the Australian House of Representatives, 1903–1906,}                                                   \\
                           & {\color[HTML]{9D0C0C} which member does a south Melbourne electorate have?}           

\\
\hline

\textit{\textbf{original}   }                &Which position had fewer rounds than 3, and an overall of less than 48?

 \\
{\textit{\textbf{de-contexualized}}} & {\color[HTML]{00009B} Which position among 2007 Jacksonville Jaguars draft history had } \\
                           &  {\color[HTML]{00009B} fewer rounds than 3, and an overall of less than 48?
 }                 \\
{\textit{\textbf{parapharsed}} }     &  {\color[HTML]{9D0C0C}Which position in the 2007 Jacksonville Jaguars draft history }                                                   \\
                           & {\color[HTML]{9D0C0C} had less than 3 rounds and less than 48 overall?} \\
                           \hline

\textit{\textbf{original}   }                &How many numbers of dances for place 1?

 \\
{\textit{\textbf{de-contexualized}}} & {\color[HTML]{00009B}How many numbers of dances for place 1 for Dancing on Ice (series 5)? } \\
                          
{\textit{\textbf{parapharsed}} }     &  {\color[HTML]{9D0C0C}How many dances for 1st place for Dancing on Ice (series 5) ?}                                                   \\

\Xhline{1.5pt} 

\end{tabular}
}
            
\begin{tabular}{ll}
\Xhline{1.3pt} 
\textit{\textbf{Error type 1 (70\%)}}       & Ambiguity in the original question \\ 
\hline
\textit{\textbf{original}}          & Name the 2005 with 2007 of sf                                             \\
\textit{\textbf{de-contextualized}} & Name the 2005 with 2007's Doubles name of sf among Alicia Molik?          \\
\textit{\textbf{parapharsed}}       & Do you name the 2005s with 2007 doubles names from sf under Alicia Molik? \\
\Xhline{1.3pt}

\textit{\textbf{Error type 2 (15\%)}}       & Change of intent after decontextualizing \\ 
\hline
    \textit{\textbf{original}}          & how many total rounds did she fight in ?                                             \\
\textit{\textbf{de-contextualized}} & How many total rounds did she fight for Gina Carano?          \\
\textit{\textbf{parapharsed}}       & How many rounds did she fight for Gina Carano in total? \\ 
     \Xhline{1.3pt} 

\textit{\textbf{Error type 3 (15\%)}}       & Change of intent after paraphrasing \\ 
\hline
       \textit{\textbf{original}}          & Which Bask has an Indoor track of 0, and a Swimming of 5?                                             \\
\textit{\textbf{de-contextualized}} & Which bask has an indoor track of 0,and a swimming of 5 for horizon league's           \\
&women's sports championship totals?\\
\textit{\textbf{parapharsed}}       & Which pool has an indoor stretch of 0 and a swim of 5 for the total number  \\
&of women's athletic championships in the horizon league?\\ \Xhline{1.3pt}  
\end{tabular}

    }
    \caption{Examples of each step in the process of creating the Open-WikiTable}
    \label{table6}
    \end{table*}

        \begin{table*}[!ht]
    \centering
    \setlength{\heavyrulewidth}{1.5pt}
    \resizebox{2.0\columnwidth}{!}{
        \newcommand{\mytable}{
\begin{tabular}{ll}
\Xhline{1.3pt} 
\textit{\textbf{Error type 1 (70\%)}}       & Ambiguity in the original question \\ 
\hline
\textit{\textbf{original}}          & Name the 2005 with 2007 of sf                                             \\
\textit{\textbf{de-contextualized}} & Name the 2005 with 2007's Doubles name of sf among Alicia Molik?          \\
\textit{\textbf{parapharsed}}       & Do you name the 2005s with 2007 doubles names from sf under Alicia Molik? \\
\Xhline{1.3pt}

\textit{\textbf{Error type 2 (15\%)}}       & Change of intent after decontextualizing \\ 
\hline
    \textit{\textbf{original}}          & how many total rounds did she fight in ?                                             \\
\textit{\textbf{de-contextualized}} & How many total rounds did she fight for Gina Carano?          \\
\textit{\textbf{parapharsed}}       & How many rounds did she fight for Gina Carano in total? \\ 
     \Xhline{1.3pt} 

\textit{\textbf{Error type 3 (15\%)}}       & Change of intent after paraphrasing \\ 
\hline
       \textit{\textbf{original}}          & Which Bask has an Indoor track of 0, and a Swimming of 5?                                             \\
\textit{\textbf{de-contextualized}} & Which bask has an indoor track of 0,and a swimming of 5 for horizon league's           \\
&women's sports championship totals?\\
\textit{\textbf{parapharsed}}       & Which pool has an indoor stretch of 0 and a swim of 5 for the total number  \\
&of women's athletic championships in the horizon league?\\ \Xhline{1.3pt}  
\end{tabular}
}
            
\begin{tabular}{ll}
\Xhline{1.3pt} 
\textit{\textbf{Error type 1 (70\%)}}       & Ambiguity in the original question \\ 
\hline
\textit{\textbf{original}}          & Name the 2005 with 2007 of sf                                             \\
\textit{\textbf{de-contextualized}} & Name the 2005 with 2007's Doubles name of sf among Alicia Molik?          \\
\textit{\textbf{parapharsed}}       & Do you name the 2005s with 2007 doubles names from sf under Alicia Molik? \\
\Xhline{1.3pt}

\textit{\textbf{Error type 2 (15\%)}}       & Change of intent after decontextualizing \\ 
\hline
    \textit{\textbf{original}}          & how many total rounds did she fight in ?                                             \\
\textit{\textbf{de-contextualized}} & How many total rounds did she fight for Gina Carano?          \\
\textit{\textbf{parapharsed}}       & How many rounds did she fight for Gina Carano in total? \\ 
     \Xhline{1.3pt} 

\textit{\textbf{Error type 3 (15\%)}}       & Change of intent after paraphrasing \\ 
\hline
       \textit{\textbf{original}}          & Which Bask has an Indoor track of 0, and a Swimming of 5?                                             \\
\textit{\textbf{de-contextualized}} & Which bask has an indoor track of 0,and a swimming of 5 for horizon league's           \\
&women's sports championship totals?\\
\textit{\textbf{parapharsed}}       & Which pool has an indoor stretch of 0 and a swim of 5 for the total number  \\
&of women's athletic championships in the horizon league?\\ \Xhline{1.3pt}  
\end{tabular}

    }
    \caption{Error analysis on the construction stage of Open-WikiTable}
    \label{table7}
    \end{table*}

\appendix

\end{document}